\def\year{2020}\relax
\documentclass[letterpaper]{article} 
\usepackage{aaai20}  
\usepackage{times}  
\usepackage{helvet} 
\usepackage{courier}  
\usepackage[hyphens]{url}  
\usepackage{graphicx} 
\usepackage{graphicx}  
\usepackage{array}
\usepackage{amsmath,bm}
\usepackage{amssymb}
\usepackage{mathtools}
\usepackage{arydshln}
\usepackage[flushleft]{threeparttable}
\DeclareMathOperator*{\argmax}{arg\,max}
\usepackage{soul}
\usepackage{paralist}
\usepackage{rotating,makecell}
\usepackage{tabularx}

\usepackage{array,multirow}

\usepackage{subfigure}

\usepackage{color}
\newcommand{\hlc}[2][yellow]{ {\sethlcolor{#1} \hl{#2}} }
\definecolor{orange}{cmyk}{0.0000,0.1137,0.1843,0.0000}
\definecolor{green}{cmyk}{0.1172,0.0000,0.0251,0.0627}
\definecolor{purple}{cmyk}{0.0976,0.0000,0.2480,0.0353}
\definecolor{pink}{cmyk}{0.0000,0.1714,0.0367,0.0392}

\newcommand{\citet}[1]{\citeauthor{#1}\shortcite{#1}}
\newcommand{\citep}{\cite}

\title{Weakly-Supervised Opinion Summarization by Leveraging External Information}
\author{Chao Zhao\qquad Snigdha Chaturvedi\\  
Department of Computer Science\\
University of North Carolina at Chapel Hill\\
zhaochaocs@gmail.com\qquad snigdha@cs.unc.edu 
}

\begin{document}

\maketitle

\begin{abstract}
Opinion summarization from online product reviews is a challenging task, which involves identifying opinions related to various aspects of the product being reviewed. While previous works require additional human effort to identify relevant aspects, we instead apply domain knowledge from external sources to automatically achieve the same goal. This work proposes \textsc{AspMem}, a generative method that contains an array of memory cells to store aspect-related knowledge. This explicit memory can help obtain a better opinion representation and infer the aspect information more precisely. We evaluate this method on both aspect identification and opinion summarization tasks. Our experiments show that \textsc{AspMem} outperforms the state-of-the-art methods even though, unlike the baselines, it does not rely on human supervision which is carefully handcrafted for the given tasks. 
\end{abstract}

\section{Introduction}
\label{sec:intro}

Opinion summarization aims to generate a concise and digestible summary of user opinions, like those from the internet sources, such as blogs, social media, e-commerce websites, etc. It is especially helpful when the large and growing number of such opinions becomes overwhelming for users to read and process \cite{kim2011comprehensive,ding2015towards}. In this work, we focus on extractive opinion summarization from online product reviews. The goal of this task is to take a collection of reviews of the target product (e.g., a television) as input and selects a subset of review excerpts as a summary. The last two boxes of Figure \ref{fig:example} show an example of user reviews of a television and a corresponding extractive summary.

\begin{figure}[t]
 \centering
 \scriptsize
 \noindent\fbox{\parbox{0.95\columnwidth}{
\textbf{Feature descriptions}:\\
$\bullet$ \hlc[orange]{ENHANCED QUALITY}: With the X1 Extreme Processor enjoy controlled contrast \& wide range of brightness\\
$\bullet$ \hlc[green]{BEYOND HIGH DEFINITION}: 4K HDTV picture offers stunning clarity \& high dynamic range color \& detail. \\
$\bullet$ \hlc[green]{PREMIUM DISPLAY}: Enjoy vibrant colors with TRILUMINOS \& clear on-screen action with X-Motion Clarity.\\
$\bullet$ \hlc[purple]{VOICE COMPATIBILITY}: 55in tv is compatible with Amazon Alexa \& Google Home to change channels \& more.
}}
\noindent\fbox{\parbox{0.95\columnwidth}{
\textbf{Review 1}:\hlc[pink]{Set up was extremely easy and the remote is simple to use.} Simply plug it in and tune to a channel. It gets 4 stars because I don't think its worth the price.\\
\textbf{Review 2}:\hlc[green]{The color and definition are excellent.} We wanted a small TV for our kitchen counter...and it fit the bill, it seemed.\\
\textbf{Review 3}: I have owned this TV for 10 months and am looking to replace it. \hlc[purple]{The sound is TERRIBLE.} \hlc[green]{The picture quality is also very rapidly decreasing.}\\
\textbf{Review n}: ...
}}
\noindent\fbox{\parbox{0.95\columnwidth}{
\textbf{Summary}: \hlc[pink]{Set up was extremely easy and the remote is simple to use.} \hlc[green]{The color and definition are excellent}. It's great for casual TV watching. \hlc[purple]{The sound is TERRIBLE.} \hlc[green]{The picture quality is also very rapidly decreasing}.
}}
\caption{An example of the extractive summary from multiple reviews. A review may express opinions about multiple aspects of the target product. These are shown in the figure as highlighted texts in different colors. }
\label{fig:example}
\end{figure}

This example illustrates that opinion summarization differs from the more general task of multi-document summarization \citep{lin2002single} in two major ways. First, while general summarization aims to retain the most important content, opinion summarization needs to cover a range of popular opinions and reflect their diversity \citep{di2014hybrid}. Second, opinion summary is more centered on the various \textit{aspects} (i.e., components, attributes, or properties) of the target product, and their corresponding sentiment polarities \citep{liu2015sentiment}. For example, highlighted sentences in Review 3 of Figure \ref{fig:example} express reviewer's negative opinions about the aspects of \textsc{Sound} and \textsc{Image}. To reflect these differences, \citet{hu2004mining} introduced a three-step pipeline to create an opinion summary by 1) mining product-related aspects and identifying sentences related to those aspects; 2) analyzing the sentiment of the identified sentences; and 3) summarizing the results. Each of these three tasks has often been addressed using supervised methods. Despite the fairly high performance, these methods require the corresponding human-annotated data. Even worse, they suffer from the inability to adapt across different domains or \textit{product categories} (e.g., televisions and backpacks have different aspects). In this paper, we address these problems without the usage of human annotation. 

Previous works addressed these problems using pure unsupervised methods, but found it is challenging to detect the aspect-related segments of reviews (e.g., those highlighted in Figure \ref{fig:example}) with both high precision and recall \cite{he2017unsupervised}. A better solution is to utilize knowledge sourced from existing external information about the target product i.e., the information beyond the customers' reviews. For example, on Amazon's product webpage, we can obtain not only customer reviews but also product-related information, such as the overall description, the feature descriptions (The top of Figure \ref{fig:example} gives an example), and attributes tables. These external information sources widely exist on e-commerce websites and are easily accessible. More importantly, they are closely related to the aspects of products and therefore are great resources to facilitate the aspect identification task. 
Automatically learning aspects from such external sources can reduce the risk that human-assigned aspects may be biased, unrepresentative, or not have the desired granularity. Meanwhile, it makes the model easy to adapt to different product categories. Here we use the feature descriptions of products as the information source, and leave other sources for future work.

In this work, we propose a generative approach that relies on the aspect-aware memory (\textsc{AspMem}) to better leverage this knowledge during aspect identification and opinion summarization. \textsc{AspMem}, which is inspired by Memory Networks \cite{weston2014memory}, is an array of memory cells to store aspect-related knowledge obtained from external information. These memory cells cooperate with the model throughout learning, and judge the relevance of review sentences to the product aspects. Then the relevance is combined with the sentiment strength to determine the salience of an opinion. Finally, we extract a subset of salient opinions to create the final summary. By formalizing the subset selection process as an Integer Linear Programming (ILP) problem, the resulting summary maximizes the collective salience scores of the selected sentences while minimizing information redundancy. 

We demonstrate the benefits of our model on two tasks: aspect identification and opinion summarization, by comparing with previous state-of-the-art methods. On the first task, we show that even without any parameters to tune, our model still outperforms previously reported results, and can be further enhanced by introducing extra trainable parameters. For the summarization task, our method exceeds baselines on a variety of evaluation measures. 

Our main contributions are three-fold:
\begin{itemize}
\item We address the task of opinion summarization without using any task-specific human supervision, by incorporating domain knowledge from external information.
\item We propose a generative approach to better leverage such knowledge.
\item We experimentally demonstrate the effectiveness of the proposed method on both aspect identification and summarization tasks. 
\end{itemize}

\section{Related Work}
\label{sec:related}
 
This work spans two lines of research: aspect identification of review text, and review summarization, which are discussed next.

\subsection{Aspect identification}
Customers give their aspect-related opinions by either explicitly mentioning the aspects (e.g., high \textit{price}) or using implicit expressions (e.g., expensive), which makes aspect identification a challenging task. Supervised methods use sequence labeling models or text classifiers to identify the aspects \cite{liu2015fine}. Rule-based methods rely on frequent noun phrases and syntactic patterns \citep{hu2004mining,raju2009unsupervised}. Most unsupervised methods are based on LDA and its variants, and interpret the latent topics in reviews as aspects \citep{mei2007topic,wang2016mining}. However, LDA does not perform well in finding coherent topics from short reviews. Also, while topics and aspects may overlap, there is no guarantee that these two are the same.

To address the first problem, \citet{he2017unsupervised} propose ABAE, an unsupervised neural architecture, to enhance the topic coherence by leveraging pre-trained word embeddings. They learn the embedding for each aspect from the word embedding space through a reconstruction loss. For the second problem, \citet{angelidis2018summarizing} propose MATE, which determines the aspect embeddings in ABAE using embeddings of a few aspect-related seed-words. These seed-words are extracted from a small dataset (about 1K sentences) with human-annotated aspect labels. We borrow their idea of using aspect embeddings and seed-words. The difference is that we collect the seed-words from external information automatically. Also, while both of their models are discriminative, we propose a generative model to better leverage the seed-words.

\subsection{Opinion summarization}
Most methods in multi-documents summarization are \textit{extractive} in nature, i.e., rank and select a subset of salient segments (i.e., words, phrases, sentences, etc.) from reviews to form a concise summary \cite{kim2011comprehensive}. The ranking of each unit relies on a score to evaluate its salience, and the selection is conducted greedily \cite{wan2007manifold} or globally \cite{mcdonald2007study,nishikawa2010opinion,cao2015ranking}. For example, \citet{yu2016product} score phrases based on their popularity and specificity. \citet{ganesan2012micropinion} rank phrases based on their representativeness and readability and then create the summary via depth-first search. \citet{angelidis2018summarizing} combine aspect and sentiment to identify salient opinions, which is also adopted in our work. The difference is that we use a more precise and flexible method to calculate the aspect-relevance of reviews. Meanwhile, rather than selecting the review segments greedily which can yield sub-optimal solutions, we use ILP to find its optimal subset. 

To the best of our knowledge, the only work that uses external information to enhance summarization is by \citet{narayan2017neural}, who use title and image captions to assist supervised news summarization. Another direction focuses on \textit{abstractive} methods to generate new sentences from the source text \cite{ganesan2010opinosis,chu2019meansum,bravzinskas2019unsupervised}.

\section{Problem Formulation}
\label{sec:overview}
Extractive opinion summarization aims to select a subset of important opinions from the entire opinion set. For product reviews, the opinion set is a collection of review segments of a certain product. Formally, we use $\mathcal{P}_{c_i}$ to denote all the products belonging to the $i$-th category $c_i$ (e.g., televisions or bags) in the corpus. Given a target product $p\in \mathcal{P}_{c_i}$, the corpus contains $m$ reviews $\mathcal{R}_p = \cup_{j=1}^m \mathcal{R}_p^{(j)}$ of this product, while each review $\mathcal{R}_p^{(j)}$ contains $n$ segments $\{s_1, s_2, \cdots, s_n\}$. We also collect the feature description $\mathcal{F}_p$ of the product as external information, which contains $\ell$ feature items $\{f_1, f_2, \cdots, f_{\ell}\}$.
The summarization model aims to select a subset of important opinions $\mathcal{O}_p \subseteq \mathcal{R}_p$ that summarize reviews of the product $p$. 

As previously mentioned, one challenge during summarization is to identify aspect-related opinions. In Sec. \ref{sec:asp}, we show how the proposed \textsc{AspMem} can tackle this problem, and how to incorporate domain knowledge to enhance model performance. The ranking and selection of the review segments are described in Sec. \ref{sec:sum}.

\section{Aspect Identification}
\label{sec:asp}
\subsection{\textsc{AspMem: }Aspect-aware memory}\label{sec:aspmem}
This section describes the proposed \textsc{AspMem} model to identify the aspect-related review segments. \textsc{AspMem} contains an array of memory cells $\mathcal{A}=\{a_1, a_2, \cdots, a_k\}$ to store aspect-related information. Each cell $a_i$ relates to one specific aspect, and has a low-dimensional embedding $\bm{a}_i\in \mathbb{R}^d$ in the semantic space, where $d$ is the dimension of the embedding. Each word $v_i$ in a review segment $s=\{v_1, v_2, \cdots, v_n\}$ also has an embedding $\bm{v}_i\in \mathbb{R}^d$ in the same semantic space.

Similar to topic models, we assume the review segment $s$ is generated from these aspect (topic) memories. However, the LDA-based topic models parameterize the generation probability at word-level, which is too flexible to model short segments in reviews \cite{yan2013biterm}. We instead regard the review segment as a whole from a single aspect during generation, but allow every word to have a different contribution to the segment representation.

Given a review segment $s$, the probability that this segment is generated by the $i$-th aspect $a_i$ is proportional to the cosine similarity of their vector representations:
\begin{equation} \label{eq:gen_prob}
 P(s|a_i) \propto \exp(\cos(\bm{s}, \bm{a}_i)),
\end{equation}
where $\bm{s}$ is the embedding of the segment $s$, and is defined as the weighted average over embeddings of the words in $s$:
\begin{equation}
 \bm{s} = \sum_i z_i \bm{v}_i.
\end{equation}
$z_i$ is the attention weight of the word $v_i$ and is proportional to $v_i$'s generation probability. That is, we focus more on those words which are more likely to be generated by the aspect memories. To compute these weights, we define the probability of $v_i$ being generated from $a_j$ in a similar way:
\begin{gather}
 P(v_i|a_j) \propto \exp(\cos(\bm{v}_i, \bm{a}_j)),\\
 P(v_i) = \sum_j P(v_i|a_j) P(a_j) \label{eq:word_prob},\\
 z_i = \frac{P(v_i)}{\sum_j P(v_j)}.
\end{gather}

Without any prior domain knowledge of the aspects, the latent embeddings $\bm{a}_j$ and the prior probabilities of aspects $P(a_j)$ are parameters (denoted by $\bm{\theta}$) and can be estimated by minimizing the negative log-likelihood of the corpus $\mathcal{X}$ (i.e., all the review segments belonging to the same product category):
\begin{equation} \label{eq:J}
 J(\theta) = -\sum_{s\in \mathcal{X}} \log P(s; \bm{\theta}) + \lambda \left\Vert \mathbf{\hat{\mathbf{A}}\hat{\mathbf{A}}^T}-\mathbf{I}\right\Vert_2.
\end{equation}

The estimation of the likelihood part $P(s; \bm{\theta})$ is similar to Eq.~\ref{eq:word_prob}. 
The second term is a regularization term, where $\hat{\mathbf{A}} \in \mathcal{R}^{k\times d}$ is the aspect embedding matrix with $\ell_2$ row normalization, and $\mathbf{I}$ is the identity matrix. It encourages the learned aspects to be diverse, i.e., the aspect embeddings are encouraged to be orthogonal to each other. $\lambda$ is the hyper-parameter of the regularization. 

Once we obtain all the parameters, we can calculate the probability of the review segment $s$ belonging to the aspect $a_i$ as
\begin{equation} \label{eq:pred}
P(a_i|s) \propto P(s|a_i)P(a_i),
\end{equation}
and then select the aspect with the highest posterior probability as the identified aspect.

\subsection{Incorporating Domain knowledge }
\label{subsec:seeds}
The aspect embeddings estimated merely from the data have several shortcomings. First, the model may learn some topics that are irrelevant to the aspects of products, such as sentiments and user profiles. Second, it is difficult to control the granularity of the learned aspects, which may lead to too coarse- or fine-grained aspects. 

To address these problems, a simple yet effective method is to use domain knowledge about products. Specifically, rather than estimating $\bm{a}_i$ according to Eq.~\ref{eq:J}, one could collect several aspect-related seed-words, (e.g., \textit{picture}, \textit{color}, \textit{resolution}, and \textit{bright} for the \textsc{Display} aspect), and average the embeddings of these seed-words to produce $\bm{a}_i$. Previous works have shown the benefit of such knowledge \cite{fast2017lexicons,angelidis2018summarizing}, but they have to encode this knowledge manually or from the human-annotated data, which makes these methods less easy to adapt across product categories. 

As we mentioned in Sec. \ref{sec:intro}, feature descriptions of products can be a valuable external resource for seed-words mining. Here we describe our unsupervised method of collecting the seed-words from it. To increase the size of this resource, we assume all products in the same category have shared aspects, and collect seed-words from the category level. For each product category $c_i$, we collect the feature items $\mathcal{F}^{c_i}$ from all products of the same category as the document, i.e., $\mathcal{F}^{c_i}=\bigcup_{p\in \mathcal{P}_{c_i}}\mathcal{F}_p$, and then apply TF-IDF to extract seed-words from it \footnote{We also tried other algorithms, but the differences were not significant.}. For TF-IDF to work, we need the seed-words to have high term frequency and the general words have high document frequency. We therefore aggregate all the items in $\mathcal{F}^{c_i}$ as one single document, and regard the remaining items belonging to other categories as individual documents to build the corpus. For example, assume we have six product categories, while each category contains ten products, and each product has ten feature descriptions. We therefore have 600 feature descriptions in total. To extract the seed-words of one category (e.g., the TV), we concatenate the 100 TV-related descriptions as one single document, while regarding the other 500 descriptions as individual documents. We then calculate the TF-IDF of each word based on these 501 documents. Finally, we select the top $K$ words with the highest TF-IDF value as seed-words of the product category $c_i$. 

\section{Summary Generation}
\label{sec:sum}
In summary generation stage, we first evaluate the salience of each opinion segment, and then select a subset of opinions which form the final summary.

\subsection{Salience of the opinion}
Following \citet{angelidis2018summarizing}, we evaluate the salience of a review segment $s$ from two perspectives: the relevance to aspects, and the sentiment strength. 

{\bf Relevance} depicts how relevant a segment is to the various aspects of the product. Since one segment may relate to more than one aspect (e.g., \textit{The color is excellent but the sound is terrible.}), we calculate relevance at the word level rather than the segment level. Recall that the relevance of a word to an aspect memory is proportional to the cosine similarity between their embeddings. We assign each word its most related aspect memory (by $\max$ operation), and calculate the relevance of the entire segment as the averaged relevance over all words (by $\sum$ operation). That is,
\begin{equation} \label{eq:rel}
 \mathbb{S}_{rel}(s) = \frac{1}{|s|} \sum_i \max_{j=\{1,\cdots,K\}} g(\cos (\bm{v}_i, \bm{a}_j)\cdot w_j).
\end{equation}

We use the $K$ seed-words extracted from Sec. \ref{subsec:seeds} as the aspect-related memory, and $w_j$ and $\bm{a}_j$ are the weight and word embedding of the $j$-th seed-word. Here the $\cos (\bm{v}_i, \bm{a}_j)$ and $w_j$ can be regarded as the unnormalized conditional and prior probabilities in Eq.~\ref{eq:word_prob}. $g(x)=x\cdot \bm{I}(x-\delta)$ is an activation function to filter the general words whose cosine similarity with any aspects is less than $\delta$. $\bm{I}(\cdot)$ is the step function. Compared with the relevance measure adopted by \citet{angelidis2018summarizing}, which uses the probability difference between the most probable aspect and the general one, our score takes a soft assignment between words and aspects, and thus allows the segment to relate to more than one aspect. Also, by regarding each seed-word as a fine-grained aspect, it does not require the seed-words to be clustered into aspects. 

{\bf Sentiment} reflects customers’ preferences regarding products and their aspects, which is helpful in decision making. Since sentiment analysis is not the major contribution of this work, we directly apply the CoreNLP \cite{socher2013recursive} and a sentiment lexicon \footnote{https://www.cs.uic.edu/\textasciitilde liub/FBS/sentiment-analysis.html\#lexicon} to get the sentiment distribution of the reviews. The sentiment distribution is then mapped onto $[0, 1]$ range as the sentiment score $\mathbb{S}_{senti}$. Sentences with stronger sentiment polarities will have higher values.

Finally, we evaluate the salience of one opinion segment by multiplying the two scores:
\begin{equation}
 \mathbb{S}_{sal}(s) = \mathbb{S}_{rel}(s) \times \mathbb{S}_{senti}(s).
\end{equation}

\subsection{Opinion selection}
An ideal summary would contain as many high-salience opinions as possible. However, care should be taken to avoid redundant information. Also, there has to be a limit on the length of the summary (i.e. no longer than $L$ words). These goals can be formalized as an ILP problem. We introduce an indicator variable $\alpha_i\in \{0,1\}$ to indicate whether to include the $i$-th segment $s_i$ in the final summary, and then find the optimal $\bm{\alpha}$ of the following objective:
\begin{equation}\
 \bm{\alpha} = \argmax_{\bm{\alpha}} \sum_i \mathbb{S}_{sal}(s_i) \alpha_i - \sum_{i,j} sim_{ij}\beta_{ij},
\end{equation}
\vspace{-12pt}
\begin{alignat}{3}
 & s.t. \quad && \alpha_i, \beta_{ij} \in \{0,1\} \quad && \forall i,j \\\label{eq:beta1}
 & \quad && \beta_{ij} \geq \alpha_i + \alpha_j - 1 \quad && \forall i,j \\\label{eq:beta2}
 & \quad && \beta_{ij} \leq \frac12 (\alpha_i + \alpha_j) \quad && \forall i,j\\\label{eq:len}
 & \quad && \textstyle\sum\nolimits_i \alpha_i l_i \leq L \quad && \forall i 
\end{alignat}
where $sim_{ij}$ is the similarity between $s_i$ and $s_j$. $\beta_{ij}$ is an auxiliary binary variable that will be $1$ iff both $\alpha_i$ and $\alpha_j$ equal to $1$, and this is guaranteed by Eq.~\ref{eq:beta1} - \ref{eq:beta2}. Eq.~\ref{eq:len} is used to restrict the length of the summary, where $l_i$ is the length of $s_i$. We solve the ILP with Gurobi \footnote{http://www.gurobi.com/}. 

\section{Experiments}

\subsection{Dataset}
We utilize \textsc{OpoSum}, a review summarization dataset provided by \citet{angelidis2018summarizing} to test the efficiency of the proposed method. This dataset contains about 350K reviews from the amazon review dataset \citep{he2016ups} under six product categories: \textit{Laptop bags}, \textit{Bluetooth headsets}, \textit{Boots}, \textit{Keyboards}, \textit{Televisions}, and \textit{Vacuums}. 
Each review sentence is split into segments using a rhetorical structure theory (RST) parser \citep{feng2012text} to reduce the granularity of opinions. The annotated corpus includes ten products from each category, and ten reviews from each product. They annotate each review segment with an aspect label and produce summaries for each product. We describe the details below: 

{\bf Aspect information.} 
Each product category has nine pre-defined aspect labels. Each segment is labeled with one or more aspects, including a \textsc{General} aspect if it does not discuss any specific one. The annotated dataset is split into two equal parts for validation and test. Based on the validation data, they extract 30 seed-words for each aspect and produce the corresponding aspect embedding as a weighted average of seed-words embeddings. 

{\bf Final summary. } For each product, the annotators create a summary by selecting a subset of salient opinions from the review segments and limiting its length to $100$ words. Each product has three referenced summaries created by different annotators, which are used only for evaluation.

Their dataset does not contain any external information. We therefore randomly collect the feature descriptions from about 100 products for each category. Table \ref{tab:side} gives a statistics about this data. \footnote{Available on \url{https://github.com/zhaochaocs/AspMem}}

\begin{table}[t!]
\small
 \centering
 \begin{tabular}{lllll}
 \hline \bf Category & \#prod & \#feature & \#token & vocab \\ \hline
 Bags & 254 & 5.1 & 9.2 & 1491 \\
 Headsets & 88 & 4.9 & 9.5 & 796\\
 Boots & 106 & 6.0 & 5.0 & 472\\
 Keyb/s & 142 & 4.8 & 10.5 & 1328\\
 TVs & 169 & 5.0 & 9.8 & 905\\
 Vaccums & 122 & 5.0 & 10.3 & 878\\ \hline
 \end{tabular}
 \caption{\label{tab:side} The statistics of the external data from six categories. The four columns are: the number of products, the average number of features per product, the average number of tokens per feature, and the entire vocabulary size.}
\end{table}

\subsection{Experiments on aspect identification}\label{subsec:asp_iden}

We first investigate the model's ability to identify aspects, which aims to label each review segment with one of the nine aspects (eight specific aspects and one \textsc{General} aspect) as labeled in the dataset. The method is described in Sec. \ref{sec:asp}. However, instead of using the seed-words obtained from external information (Sec. \ref{subsec:seeds}), we still use those provided with the dataset to enable fair comparison with prior works. Our external seed-words will be used in the summarization experiments (Sec. \ref{subsec:exp_sum}). 

\subsubsection{Setup} For the eight specific aspects, we assign their corresponding memory cells $\bm{a}_i$ with the average embedding of the 30 seed-words provided by \textsc{OpoSum}. For the general aspect, although \textsc{OpoSum} also provides 30 corresponding seed-words, we handle it differently for the following reasons. First, while the knowledge of specific aspects can be encoded as a few seed-words, it is hard to represent the \textsc{General} aspect in the same way. A better method is to allow the model to find its intrinsic patterns by relaxing the corresponding \textsc{General} embedding as trainable parameters. Also, since the number of the \textsc{General} reviews is approximately ten times more than the specific aspect on average, it is reasonable to assign more memory cells for the \textsc{General} aspects. Therefore, besides the fixed \textsc{General} embedding provided by MATE, we have another enhanced model with five extra memory cells to encode the \textsc{General} aspect. These extra memory cells are initialized randomly and trained to minimize the log-likelihood in Eq.~\ref{eq:J}.

We use $200$-dimensional word embeddings which are pre-trained on the training set via word2vec \citep{mikolov2013distributed}. These embeddings are fixed during training. For simplicity, the prior distribution of aspects is set as uniform. We train the model with batch size of 300, and optimize the objective using Adam \citep{kingma2014adam} with a fixed learning rate of $0.001$ and an early stopping on the development set. The $\lambda$ is set as $100$. Notice that the model without the extra aspect memories does not have any trainable parameters and therefore can directly be applied for prediction using Eq.~\ref{eq:pred}. 

We compare the proposed method with ABAE and MATE, two state-of-the-art neural methods mentioned in Sec. \ref{sec:related}, as well as a distillation approach \cite{karamanolakis2019training} that uses the pre-trained BERT \cite{devlin2019bert} as the student model.
To ensure a fair comparison, all models utilize the same seed-words. The performance is evaluated through multi-label $F_1$ score.

\begin{table*}[t]
\small
\centering
\begin{tabular}{llllllll}
\hline \bf Model & Bags & Headsets & Boots & Keyb/s & TVs & Vaccums & Average \\ \hline
ABAE \small{\cite{he2017unsupervised}} & 41.6 & 48.5 & 41.0 & 41.3 & 45.7 & 40.6 & 43.2\\
MATE \small{\cite{angelidis2018summarizing}} & 48.6 & 54.5 & 46.4 & 45.3 & 51.8 & 47.7 & 49.1\\
BERT \small{\cite{karamanolakis2019training}} & \textbf{61.4} & \textbf{66.5} & 52.0 & 57.5 & \textbf{63.0} & 60.4 & \textbf{60.2}\\
\cdashline{1-8}
\textsc{AspMem} & 52.4 & 58.1 & 54.5 & 51.4 & 53.9 & 54.6 & 54.2\\
\quad w/ extra memory & 60.0 & 62.0 & \textbf{55.8} & \textbf{61.8} & 60.0 & \textbf{61.8} & \textbf{60.2}\\
\hline
\end{tabular}
\caption{\label{tab:f1} Evaluation of the aspect identification task via multi-class $F_1$ measure. Our method outperforms MATE on all the categories and achieves a 5.1\% increase on average. The extra latent aspect embeddings for the \textsc{General} aspects further boost the performance by 6.0\%. }
\end{table*}

\subsubsection{Results} Table \ref{tab:f1} shows the average $F_1$ scores for the four models on the six categories. MATE performs better than ABAE by introducing the human-provided seed-words, which demonstrates the effectiveness of domain knowledge. However, MATE applies the same neural architecture as ABAE, which may not be the best fit to fully leverage the power of the introduced knowledge. Our generative model instead directly cooperates with the aspect memory, not only during the prediction stage but also during the segment encoding. Without any trainable parameters, our method outperforms ABAE and MATE on all the categories and achieves a 5.1\% increase on average. It indicates that \textsc{AspMem} can get a better aspect-aware segment representation for aspect identification. The extra latent aspect embeddings of the \textsc{General} aspect (\textsc{AspMem} w/ extra memory) help the model better fit the intrinsic structure of the data, which further improves the performance by 6.0\%. When comparing with BERT, our model still has better performance on three categories and achieves the same average $F_1$ score. Note that while BERT is a pre-trained model with 110M parameters, our model only has 1K parameters.

\subsubsection{Discussion} 
\begin{figure}[t]
 \centering
\includegraphics[width=0.95\columnwidth]{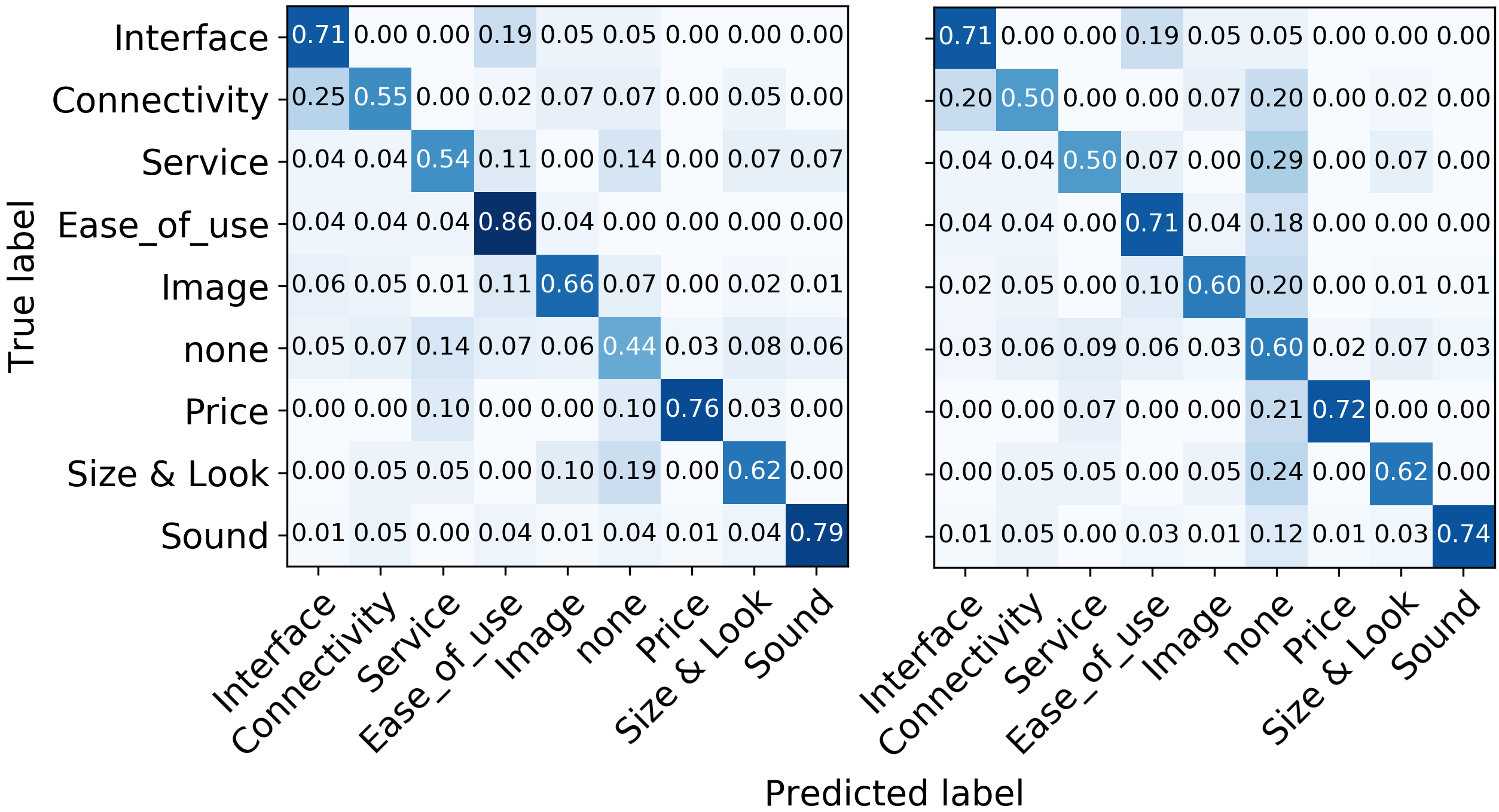}
\caption{Confusion matrix of AspMem results w/o extra memory (left) and w/ extra memory (right). Having extra memories improves performance on the \textsc{General} aspect without hurting other aspects by much.}
\label{fig:cm}
\end{figure}

To further demonstrate the contribution of the extra memories, Figure \ref{fig:cm} provides the confusion matrices of the results with and without them. The comparison shows that extra memories improve the true-positive rate of the \textsc{General} aspect from 0.44 to 0.60, while only slightly hurting those of other aspects. Table \ref{tab:seeds} shows the automatically learned \textsc{General} aspects by listing their nearest words in the embedding space. Compared with the single \textsc{General} aspect provided by MATE, our model successfully identifies the more varied \textsc{General} aspects from the reviews, such as the \textsc{Noun}, \textsc{Verb}, \textsc{Adjective}, \textsc{Number}, and \textsc{Problem}. 

\begin{table}[t!]
\footnotesize
\centering
\noindent\begin{tabular}{l|p{2in}l}
\hline Aspect & Seed-words \\ \hline
noun & tv television set hdtv item tvs product\\
adj & good great better awesome superb\\
verb & figure afford get see find hear watch\\
number & dd dddd d ddd \\
problem & issue problem occur encounter flaw \\\hline
MATE & buy purchase money sale deal week \\
\hline
\end{tabular}
\caption{\label{tab:seeds} The extra \textsc{General} aspects learned from the data, and the one provided by MATE. Numbers are delexicalized with their shape.}
\end{table}

\subsection{Experiments on Summarization}\label{subsec:exp_sum}
In this experiment, we investigate the utility of \textsc{AspMem} for summarization, using the seed-words from external sources and the selection procedure described in Sec. \ref{sec:sum}. We refer to our method as \textsc{AspMemSum}.

\subsubsection{Setup} With the method described in Sec. \ref{subsec:seeds}, we select top $100$ seed-words according to their TF-IDF values, and use their word embeddings as the $100$ aspect memories. The similarity threshold $\delta$ is set as $0.3$. The length of the summary is limited to $100$ words or less to enable comparison with the ground-truth summaries.
Similar to previous works, we add a redundancy filter to remove the repeated opinions by setting $sim_{ij}=\infty$ when $\cos(s_i, s_j) > 0.5$ otherwise as $0$. Other settings are the same as those in the last experiment. We employ ROUGE \cite{lin2004rouge} to evaluate the results. It measures the overlapping percentage of unigrams (ROUGE-1) and bigrams (ROUGE-2) between the generated and the referenced summaries. We compare our method with the reported results in \citet{angelidis2018summarizing}. 

\begin{table}[t!]
\small\centering
\begin{tabular}{lll}
\hline \bf Methods & \bf R-1 & \bf R-2 \\ \hline
Lead & 35.5 & 15.2 \\
LexRank & 37.7 & 14.1 \\
Opinosis & 36.8 & 14.3 \\ 
MATE + MILNET & 44.1 & 21.8 \\ \cdashline{1-3}
\textsc{AspMemSum} & 46.6 & 25.7 \\
\quad w/o filtering & \textbf{48.0} & \textbf{28.7} \\
\quad w/o Relevance & 41.5 & 20.5 \\
\quad w/o Sentiment & 40.5 & 18.2 \\
\quad w/o ILP & 46.2 & 25.1 \\\cdashline{1-3}
Inter-annotator Agreement & 54.7 & 36.6\\
\hline
\end{tabular}
\caption{Summarization results evaluated by Rouge. The proposed \textsc{AspMemSum} without redundancy filtering achieves the best performance on automatic metrics, and both two perform better than all the baselines.}
\label{tab:rouge}
\end{table}

\begin{table*}[t]
  \centering
  \scriptsize
  \begin{tabular}{lp{15.5cm}}
    \hline
     MATE
     & Picture is crisp and clear with lots of options to change for personal preferences. Plenty of ports and settings to satisfy most everyone. The sound is good and strong. But the numbers of options available in the on-line area of the Tv are numerous and extremely useful! I am very disappointed with this TV for two reasons : picture brightness and channel menu. The software and apps built into this TV are difficult to use and setup Unit developed a high pitch whine\\ \hline
     \textsc{AspMem}
     & Unit developed a high pitch whine. The picture is beautiful. This TV looks very good. The sound is clear as well. there is a dedicated button on the remote. I am very disappointed with this TV for two reasons : picture brightness and channel menu. which is TOO SLOW to stream HD video... and it will not work with an HDMI connection because of a conflict with Comcast's DHCP. \\ \hline
    \multirow{1}{*}{Human}
    & Picture is crisp and clear with lots of options to change for personal preferences. Plenty of ports and settings to satisfy most everyone. The sound is good and strong. But the numbers of options available in the on-line area of the Tv are numerous and extremely useful! I am very disappointed with this TV for two reasons : picture brightness and channel menu. The software and apps built into this TV are difficult to use and setup Unit developed a high pitch whine \\ \hline
  \end{tabular}
\caption{\label{tab:sum} A summary example generated by MATE and our method, compared with a human-generated summary. We use the same product (Sony BRAVIA HDTV) reported by \citet{angelidis2018summarizing}.}
\end{table*} 

\subsubsection{Results} Table \ref{tab:rouge} reports the ROUGE-1 and ROUGE-2 scores of each system \footnote{MILNET is a sentiment analyzer but its pre-trained model is not public. We therefore replaced it with CoreNLP and obtained the results of MATE as $43.9$ and $22.0$. There is no significant difference.} and the inter-annotator agreement among three annotators. Our method (\textsc{AspMemSum}) significantly outperforms the baselines on both ROUGE scores (approximate randomization \cite{noreen1989computer,chinchor1992statistical}, $N=9999, p<0.001$). When removing the redundancy filtering (w/o filtering), it achieves the highest performance. 
This observation is different from that made by \citet{angelidis2018summarizing} who found that redundancy filtering improved the ROUGE scores of results produced by MATE. 
Upon eyeballing the generated summaries we found that in absence of redundancy filtering, \textsc{AspMem}'s summaries often included the overlapping part of the three references (i.e., the segments with similar opinions but from different references) more than once. This results in the improvement of ROUGE scores: the more matched n-grams are found, the better the results. However, we prefer to avoid redundancy in order to improve readability. 

\subsubsection{Effectiveness of opinion selection}
During the opinion selection, we conduct an ablation study to investigate the contribution of the two salience scores:  $\mathbb{S}_{rel}(s)$ for the relevance and $\mathbb{S}_{senti}(s)$ for the sentiment. As shown in Table \ref{tab:rouge}, removing the relevance score drops R1 and R2 by 5.1 and 5.2, respectively. Similarly, without sentiment, R1 and R2 drop by 6.1 and 7.5. It demonstrates that both these scores are necessary to capture the salience of an opinion segment.

Finally, we back off our opinion selection procedure to the greedy method to have a fairer comparison with the baseline. As shown in Table \ref{tab:rouge} (w/o ILP), under the same greedy strategy, our method still outperforms the baselines, but using ILP can further improve the results. 

\subsubsection{Effectiveness of seed-words}
\begin{figure}[t]
 \centering
\includegraphics[width=0.8\columnwidth]{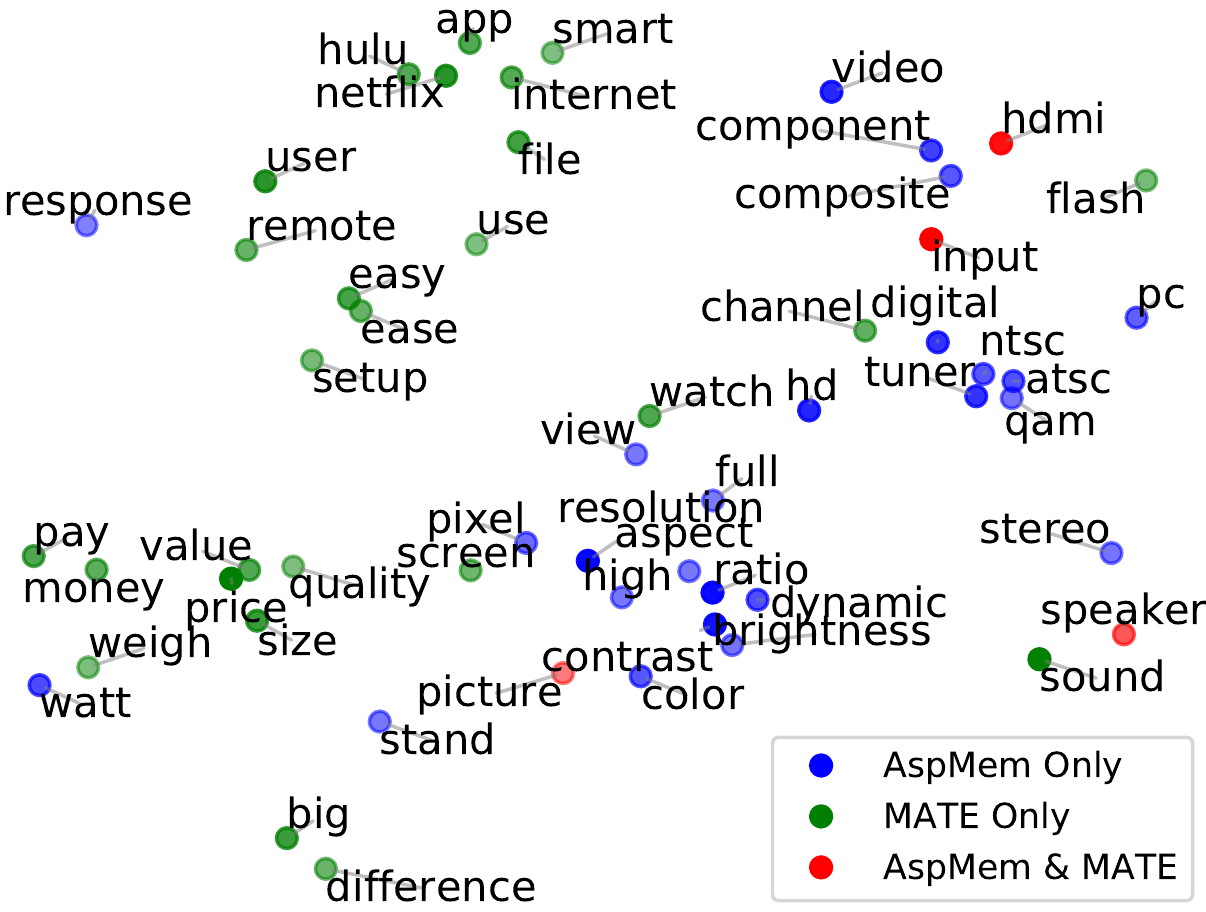}
\caption{\label{fig:seeds} The distribution of seed-words in embedding space through t-SNE \cite{maaten2008visualizing}. Each node represents a seed-word and is colored according to the seed-sets it belongs to. Words with higher weights have higher degree of opacity.}
\end{figure}

During the summarization, we extract the seed-words $\mathcal{V}_1$ from external information, whereas those used in MATE (denote by $\mathcal{V}_2$) are extracted from customer reviews with the help of aspect labels. Figure \ref{fig:seeds} provide the distribution of two seed-sets in word embedding space. We analyzed the difference between the two seed-sets, and find that about $81\%$ of words in one seed-set do not appear in the other seed-set. Even the remaining $19\%$ shared seed-words have different weights. Another observation is that the seed-words from feature descriptions tend to be nouns, while those from review texts contain more adjectives. It can also be reflected in Figure \ref{fig:seeds}, where the words from two seed-sets are separated into two parts. It reflects the fact that the content in feature descriptions is more objective than that in customer reviews, making it a better source to analyze the aspect relevancy than the reviews themselves. 

We then replace our seed-words with those used in MATE to delineate the contributions of the model from that of the seed-set. When using the same seed-words, our model achieves 45.6 and 24.5 for ROUGE-1 and ROUGE-2, which are still better than the results of MATE. This indicates that the model itself also contributes to the performance gain. 

\begin{figure}[t]
 \centering
\includegraphics[width=0.95\columnwidth]{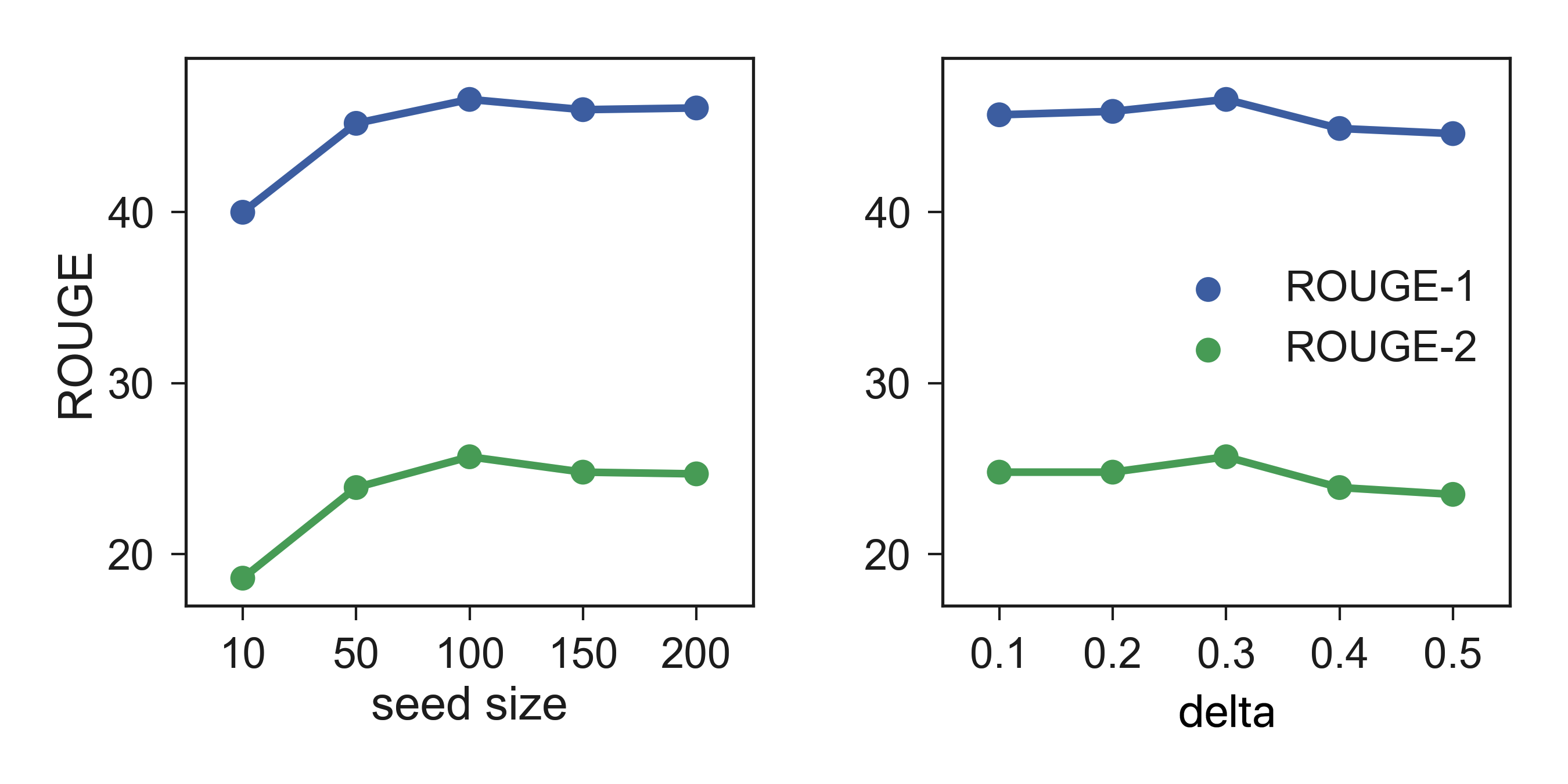}
\caption{The effect of the seeds size (left) and the similarity threshold (right) on the ROUGE metrics.}
\label{fig:seed_size}
\end{figure}

Finally, we analyze the effect of two seeds-related hyperparameters on ROUGE metrics: the size of the seed-set, and the similarity threshold $\delta$ of seed-words (see $g(\cdot)$ in Eq. \ref{eq:rel}). We vary the size of the seed-set from 10 to 200, and $\delta$ from 0.1 to 0.5. The results are shown in Figure \ref{fig:seed_size}. When there are only a few seed-words, the model performance rapidly increases with the growth of the seed-set size. For larger seed-sets (more than $100$ words), the number of noisy words increases and this slightly hurts the performance. Meanwhile, we find that our model is also robust to the choice of $\delta$, especially for small values (less than $0.3$). 

\subsubsection{Qualitative analysis}
Table \ref{tab:sum} shows summaries of the same product generated by MATE, our method (\textsc{AspMemSum}), and one of the human annotators. Similar to humans, MATE and \textsc{AspMemSum} are also able to select aspect-related opinions. The difference is that \textsc{AspMemSum} learns these aspects without any human effort.

\section{Conclusion}
In this work, we propose a generative approach to create summaries from online product reviews without specific human annotation. At the model level, we introduce the aspect-aware memory to fully leverage the domain knowledge. It also reduces the parameters and computation cost of the model. At the data level, we collect the domain knowledge from external information rather than through human effort, which makes the proposed method easier to adapt to other product categories. By comparing with the state-of-the-art models on both aspect identification and opinion summarization tasks, we experimentally demonstrate the effectiveness of our approach. Future works can design better measures for opinion selection, and incorporate abstractive methods to enhance readability of the generated summaries.

\fontsize{9.0pt}{10.0pt}
\bibliography{aaai_short}

\begin{thebibliography}{}

\bibitem[\protect\citeauthoryear{Angelidis and
  Lapata}{2018}]{angelidis2018summarizing}
Angelidis, S., and Lapata, M.
\newblock 2018.
\newblock Summarizing opinions: Aspect extraction meets sentiment prediction
  and they are both weakly supervised.
\newblock In {\em Proceedings of the 2018 Conference on EMNLP},  3675--3686.

\bibitem[\protect\citeauthoryear{Bra{\v{z}}inskas, Lapata, and
  Titov}{2019}]{bravzinskas2019unsupervised}
Bra{\v{z}}inskas, A.; Lapata, M.; and Titov, I.
\newblock 2019.
\newblock Unsupervised multi-document opinion summarization as copycat-review
  generation.
\newblock {\em arXiv preprint arXiv:1911.02247}.

\bibitem[\protect\citeauthoryear{Cao \bgroup et al\mbox.\egroup
  }{2015}]{cao2015ranking}
Cao, Z.; Wei, F.; Dong, L.; Li, S.; and Zhou, M.
\newblock 2015.
\newblock Ranking with recursive neural networks and its application to
  multi-document summarization.
\newblock In {\em 29th AAAI conference}.

\bibitem[\protect\citeauthoryear{Chinchor}{1992}]{chinchor1992statistical}
Chinchor, N.
\newblock 1992.
\newblock The statistical significance of the muc-4 results.
\newblock In {\em Proceedings of the 4th MUC},  30--50.
\newblock ACL.

\bibitem[\protect\citeauthoryear{Chu and Liu}{2019}]{chu2019meansum}
Chu, E., and Liu, P.
\newblock 2019.
\newblock Meansum: a neural model for unsupervised multi-document abstractive
  summarization.
\newblock In {\em ICML},  1223--1232.

\bibitem[\protect\citeauthoryear{Devlin \bgroup et al\mbox.\egroup
  }{2019}]{devlin2019bert}
Devlin, J.; Chang, M.-W.; Lee, K.; and Toutanova, K.
\newblock 2019.
\newblock Bert: Pre-training of deep bidirectional transformers for language
  understanding.
\newblock In {\em Proceedings of the 2019 Conference of NAACL-HLT},
  4171--4186.

\bibitem[\protect\citeauthoryear{Di~Fabbrizio, Stent, and
  Gaizauskas}{2014}]{di2014hybrid}
Di~Fabbrizio, G.; Stent, A.; and Gaizauskas, R.
\newblock 2014.
\newblock A hybrid approach to multi-document summarization of opinions in
  reviews.
\newblock In {\em Proceedings of the 8th INLG Conference},  54--63.

\bibitem[\protect\citeauthoryear{Ding and Jiang}{2015}]{ding2015towards}
Ding, Y., and Jiang, J.
\newblock 2015.
\newblock Towards opinion summarization from online forums.
\newblock In {\em Proceedings of RANLP},  138--146.

\bibitem[\protect\citeauthoryear{Fast, Chen, and
  Bernstein}{2017}]{fast2017lexicons}
Fast, E.; Chen, B.; and Bernstein, M.~S.
\newblock 2017.
\newblock Lexicons on demand: Neural word embeddings for large-scale text
  analysis.
\newblock In {\em IJCAI},  4836--4840.

\bibitem[\protect\citeauthoryear{Feng and Hirst}{2012}]{feng2012text}
Feng, V.~W., and Hirst, G.
\newblock 2012.
\newblock Text-level discourse parsing with rich linguistic features.
\newblock In {\em Proceedings of the 50th ACL},  60--68.

\bibitem[\protect\citeauthoryear{Ganesan, Zhai, and
  Han}{2010}]{ganesan2010opinosis}
Ganesan, K.; Zhai, C.; and Han, J.
\newblock 2010.
\newblock Opinosis: A graph based approach to abstractive summarization of
  highly redundant opinions.
\newblock In {\em Proceedings of Coling 2010},  340--348.

\bibitem[\protect\citeauthoryear{Ganesan, Zhai, and
  Viegas}{2012}]{ganesan2012micropinion}
Ganesan, K.; Zhai, C.; and Viegas, E.
\newblock 2012.
\newblock Micropinion generation: an unsupervised approach to generating
  ultra-concise summaries of opinions.
\newblock In {\em Proceedings of the 21st international conference on WWW},
  869--878.
\newblock ACM.

\bibitem[\protect\citeauthoryear{He and McAuley}{2016}]{he2016ups}
He, R., and McAuley, J.
\newblock 2016.
\newblock Ups and downs: Modeling the visual evolution of fashion trends with
  one-class collaborative filtering.
\newblock In {\em proceedings of the 25th international conference on WWW},
  507--517.

\bibitem[\protect\citeauthoryear{He \bgroup et al\mbox.\egroup
  }{2017}]{he2017unsupervised}
He, R.; Lee, W.~S.; Ng, H.~T.; and Dahlmeier, D.
\newblock 2017.
\newblock An unsupervised neural attention model for aspect extraction.
\newblock In {\em Proceedings of the 55th ACL},  388--397.

\bibitem[\protect\citeauthoryear{Hu and Liu}{2004}]{hu2004mining}
Hu, M., and Liu, B.
\newblock 2004.
\newblock Mining and summarizing customer reviews.
\newblock In {\em Proceedings of the tenth ACM SIGKDD international conference
  on KDD},  168--177.
\newblock ACM.

\bibitem[\protect\citeauthoryear{Karamanolakis, Hsu, and
  Gravano}{2019}]{karamanolakis2019training}
Karamanolakis, G.; Hsu, D.; and Gravano, L.
\newblock 2019.
\newblock Training neural networks for aspect extraction using descriptive
  keywords only.
\newblock In {\em The 2nd Learning from Limited Labeled Data (LLD) Workshop}.

\bibitem[\protect\citeauthoryear{Kim \bgroup et al\mbox.\egroup
  }{2011}]{kim2011comprehensive}
Kim, H.~D.; Ganesan, K.; Sondhi, P.; and Zhai, C.
\newblock 2011.
\newblock Comprehensive review of opinion summarization.
\newblock Technical report, UIUC.

\bibitem[\protect\citeauthoryear{Kingma and Ba}{2014}]{kingma2014adam}
Kingma, D.~P., and Ba, J.
\newblock 2014.
\newblock Adam: A method for stochastic optimization.
\newblock {\em arXiv:1412.6980}.

\bibitem[\protect\citeauthoryear{Lin and Hovy}{2002}]{lin2002single}
Lin, C.-Y., and Hovy, E.
\newblock 2002.
\newblock From single to multi-document summarization.
\newblock In {\em Proceedings of the 40th ACL}.

\bibitem[\protect\citeauthoryear{Lin}{2004}]{lin2004rouge}
Lin, C.-Y.
\newblock 2004.
\newblock Rouge: A package for automatic evaluation of summaries.
\newblock {\em Text Summarization Branches Out}.

\bibitem[\protect\citeauthoryear{Liu, Joty, and Meng}{2015}]{liu2015fine}
Liu, P.; Joty, S.; and Meng, H.
\newblock 2015.
\newblock Fine-grained opinion mining with recurrent neural networks and word
  embeddings.
\newblock In {\em Proceedings of the 2015 Conference on EMNLP},  1433--1443.

\bibitem[\protect\citeauthoryear{Liu}{2015}]{liu2015sentiment}
Liu, B.
\newblock 2015.
\newblock {\em Sentiment analysis: Mining opinions, sentiments, and emotions}.
\newblock Cambridge University Press.

\bibitem[\protect\citeauthoryear{Maaten and
  Hinton}{2008}]{maaten2008visualizing}
Maaten, L. v.~d., and Hinton, G.
\newblock 2008.
\newblock Visualizing data using t-sne.
\newblock {\em JMLR} 9(Nov):2579--2605.

\bibitem[\protect\citeauthoryear{McDonald}{2007}]{mcdonald2007study}
McDonald, R.
\newblock 2007.
\newblock A study of global inference algorithms in multi-document
  summarization.
\newblock In {\em European Conference on Information Retrieval},  557--564.
\newblock Springer.

\bibitem[\protect\citeauthoryear{Mei \bgroup et al\mbox.\egroup
  }{2007}]{mei2007topic}
Mei, Q.; Ling, X.; Wondra, M.; Su, H.; and Zhai, C.
\newblock 2007.
\newblock Topic sentiment mixture: modeling facets and opinions in weblogs.
\newblock In {\em Proceedings of the 16th international conference on WWW},
  171--180.
\newblock ACM.

\bibitem[\protect\citeauthoryear{Mikolov \bgroup et al\mbox.\egroup
  }{2013}]{mikolov2013distributed}
Mikolov, T.; Sutskever, I.; Chen, K.; Corrado, G.~S.; and Dean, J.
\newblock 2013.
\newblock Distributed representations of words and phrases and their
  compositionality.
\newblock In {\em NIPS},  3111--3119.

\bibitem[\protect\citeauthoryear{Narayan \bgroup et al\mbox.\egroup
  }{2017}]{narayan2017neural}
Narayan, S.; Papasarantopoulos, N.; Cohen, S.~B.; and Lapata, M.
\newblock 2017.
\newblock Neural extractive summarization with side information.
\newblock {\em arXiv:1704.04530}.

\bibitem[\protect\citeauthoryear{Nishikawa \bgroup et al\mbox.\egroup
  }{2010}]{nishikawa2010opinion}
Nishikawa, H.; Hasegawa, T.; Matsuo, Y.; and Kikui, G.
\newblock 2010.
\newblock Opinion summarization with integer linear programming formulation for
  sentence extraction and ordering.
\newblock In {\em Proceedings of the 23rd ICCL: Posters},  910--918.
\newblock ACL.

\bibitem[\protect\citeauthoryear{Noreen}{1989}]{noreen1989computer}
Noreen, E.~W.
\newblock 1989.
\newblock {\em Computer-intensive methods for testing hypotheses}.
\newblock Wiley New York.

\bibitem[\protect\citeauthoryear{Raju, Pingali, and
  Varma}{2009}]{raju2009unsupervised}
Raju, S.; Pingali, P.; and Varma, V.
\newblock 2009.
\newblock An unsupervised approach to product attribute extraction.
\newblock In {\em European Conference on Information Retrieval},  796--800.
\newblock Springer.

\bibitem[\protect\citeauthoryear{Socher \bgroup et al\mbox.\egroup
  }{2013}]{socher2013recursive}
Socher, R.; Perelygin, A.; Wu, J.; Chuang, J.; Manning, C.~D.; Ng, A.; and
  Potts, C.
\newblock 2013.
\newblock Recursive deep models for semantic compositionality over a sentiment
  treebank.
\newblock In {\em Proceedings of the 2013 conference on EMNLP},  1631--1642.

\bibitem[\protect\citeauthoryear{Wan, Yang, and Xiao}{2007}]{wan2007manifold}
Wan, X.; Yang, J.; and Xiao, J.
\newblock 2007.
\newblock Manifold-ranking based topic-focused multi-document summarization.
\newblock In {\em IJCAI}, volume~7,  2903--2908.

\bibitem[\protect\citeauthoryear{Wang, Chen, and Liu}{2016}]{wang2016mining}
Wang, S.; Chen, Z.; and Liu, B.
\newblock 2016.
\newblock Mining aspect-specific opinion using a holistic lifelong topic model.
\newblock In {\em Proceedings of the 25th international conference on WWW},
  167--176.

\bibitem[\protect\citeauthoryear{Weston, Chopra, and
  Bordes}{2014}]{weston2014memory}
Weston, J.; Chopra, S.; and Bordes, A.
\newblock 2014.
\newblock Memory networks.
\newblock {\em arXiv:1410.3916}.

\bibitem[\protect\citeauthoryear{Yan \bgroup et al\mbox.\egroup
  }{2013}]{yan2013biterm}
Yan, X.; Guo, J.; Lan, Y.; and Cheng, X.
\newblock 2013.
\newblock A biterm topic model for short texts.
\newblock In {\em Proceedings of the 22nd international conference on WWW},
  1445--1456.
\newblock ACM.

\bibitem[\protect\citeauthoryear{Yu \bgroup et al\mbox.\egroup
  }{2016}]{yu2016product}
Yu, N.; Huang, M.; Shi, Y.; et~al.
\newblock 2016.
\newblock Product review summarization by exploiting phrase properties.
\newblock In {\em Proceedings of COLING 2016},  1113--1124.

\end{thebibliography}
\bibliographystyle{aaai}
\end{document}